# Dutch Cross Serial Dependencies in HPSG

Gerrit Rentier*
Institute for Language Technology and Artificial Intelligence
Tilburg University, PO Box 90153, 5000 LE Tilburg, The Netherlands
rentier@kub.nl

**Abstract**

We present an analysis of Dutch cross serial dependencies in Head-driven Phrase Structure Grammar ([P&S(1994)]). We start out from the assumption that causative and perceptual verbs, like auxiliaries, can lexically 'raise' the arguments of the non-finite verbs they govern to their own list of arguments through "*argument composition*" ([H&N(1989)]).

## 1 Introduction

Dutch cross serial dependencies (DCSDs), well-known from (1) and (2), still challenge computational linguistics for an efficient treatment.

(1) dat ik$_1$ haar$_2$ de nijlpaarden$_2$ zag$_1$ voeren$_2$
    that I  her   the hippos      saw   feed

   "that I saw her feed the hippos"

(2) dat ik$_1$ Henk$_2$ haar$_3$ de nijlpaarden$_3$ zag$_1$
    that I  Henk   her    the hippos      saw
    helpen$_2$ voeren$_3$
    help    feed

   "that I saw Henk help her feed the hippos"

The problematic aspects of DCSDs are of course the bounded *discontinuous* relation between the NPs and the verbs of which they are arguments, indicated in (1) and (2) by the subscripted integers, and the *recursiveness* of the phenomenon. The construction is only licensed by members of two closed classes of verbs, the class of perceptual verbs like *zien*("see"), *horen*("hear") and *voelen*("feel"), and the class of causative verbs like *laten*("let/make") and *helpen*("help"). In the analysis put forward here we emphasize this lexical aspect of the phenomenon; in our analysis DCSDs are strictly tied to the subcategorization and semantics of perceptual and causative verbs. We analyze them as verbs which select, apart from their subject, a nonfinite V-projection which denotes an event. More particularly, as is proposed for German auxiliaries in [H&N(1989)], they subcategorize for the arguments of the verb they govern, a mechanism frequently referred to as *argument composition* or argument *inheritance*.

Recently DCSDs have been analyzed in a non-standard version of HPSG[1] in [Reape(fc.)]. In his so-called *sequence union* approach, the standard concept of phrase structure grammar (i.e. that a string is defined as the terminal yield of a phrase structure tree) is abandoned. Our analysis is more standard, in the sense that we only need to refer to the lexicon and the HPSG-mechanism of *structure sharing*.[2] Our preferred explanatory mechanism, argument composition, is not so much an additional mechanism as an *effect* which derives from careful specification of structure sharing, and structure sharing is already at the theoretical core of HPSG.

Furthermore, argument composition is independently motivated, because Dutch is like German with respect to the phrase-structural behaviour of auxiliaries, and argument composition in German constructions with auxiliaries is well-motivated ([H&N(1989)]). So we have good reason to assume argument composition present in the theory, regardless of DCSDs.

## 2 Event Semantics in HPSG

The choice of semantics in terms of a theory of events, known from [Davidson(1967)], offers interesting advantages and explanations of logical and linguistic phenomena, motivating the development of a constraint-based version of it.[3] So, in the spirit of event semantics we propose that main verbs like *voeren*("feed") in (3) should denote a discourse referent, which is in fact a very natural assumption. In (3) and throughout the paper, recurring $\boxed{i}$'s indicate structure sharing, that is token-identity of information, as is common usage in HPSG. Note also that we follow [Borsley(1987)] in representing subjects as values of SUBJ and follow [P&S(1994)] (chapter 9) in representing non-subject

---

*Sponsored by EC projects ESPRIT P5254 (PLUS), ESPRIT P6665 (DANDELION) and two travel grants by the Netherlands Organization for Scientific Research (NWO). Many thanks to Bob Borsley, Jo Calder, Bart Geurts, Josée Heemskerk, John Nerbonne, Paola Monachesi, Ivan Sag, Wietske Sijtsma and Craig Thiersch for detailed comments and sound advice. Errors are, of course, completely my own.

[1] Head-driven Phrase Structure Grammar, which the reader is presumed to be more or less familiar with (see [P&S(1994)]).

[2] In fact our analysis differs from many previous analyses of DCSDs in that we do *not* refer to any 'additional' (often powerful) mechanisms (sequence union, head wrapping, repeated rightward head movement).

[3] The combination of HPSG with (shallow) event semantics and translation to an event-based logical form originates with work on the EC-sponsored PLUS project, ESPRIT P5254, a *"Pragmatics-based Language Understanding System"*.

arguments as values of COMPS.

(3)
$$\begin{bmatrix} \text{PHON} & \langle \textit{voeren} \rangle \\ \text{HEAD} & \begin{bmatrix} \text{MAJOR} & \text{v} \\ \text{VFORM} & \text{base} \end{bmatrix} \\ \text{SUBJ} & \langle \text{NP}[\text{CASE}:\boxed{2}] \rangle \\ \text{COMPS} & \langle \text{NP}[\text{ACC}:\boxed{3}] \rangle \\ \text{GOV} & \langle \rangle \\ \text{CONT} & \begin{bmatrix} \text{DET} & \text{EVENT} \\ \text{PARA} & \boxed{1} \\ \text{RESTR} & \left\{ \begin{bmatrix} \text{RELN} & \text{FEED} \\ \text{INST} & \boxed{1} \\ \text{ARG1} & \boxed{2} \\ \text{ARG2} & \boxed{3} \end{bmatrix} \right\} \end{bmatrix} \\ \text{LEX} & + \end{bmatrix}$$

The constraint-based event semantics of the base form verb *voeren* as it is depicted in (3), with the quasi-determiner EVENT, should be interpreted as an existentially quantified event with a parameter $\boxed{1}$ which is restricted to involve a relation of *feeding*, an argument with the role of agent which is associated[4] with a semantic content $\boxed{2}$ and an argument associated with a semantic content $\boxed{3}$ which is the theme.[5]

Here the value of DET is a 'shallow' representation of a quantifier,[6] and the value of PARA, which is an abbreviation for 'parameter', is structure shared with the value of a feature INST which is short for 'instance'. We will suppose that the value of PARA corresponds with a discourse referent in the discourse representation associated with a natural language expression, without formally defining this relation here. The value of RESTR, which abbreviates 'restrictions', is a set of constraints on the value of this parameter.

## 3 An Argument Composition Analysis

We assume that the clause structure of DCSDs is one where we have a binary left-branching verbal complex. This verbal complex then locally selects the sum of the arguments of the verbs which constitute it. We feel that a binary branching analysis is empirically motivated by *auxiliary flip* in the same way as auxiliary flip motivates a binary right-branching structure for the German verbal complex, following [H&N(1989)].

---

[4] Here and throughout the paper, "Φ:Ψ" means "feature structure Φ with as CONTENT-value Ψ".

[5] We assume that our constraint-based event semantics is inductively translated to a level of underspecified logical form, and that this ULF-level then can be mapped to a level of logical form and a model-theoretic interpretation. The auxiliary levels are not defined here, but cf. [Rentier(ms.)].

[6] The concept of semantics we will outline here will be shallow for instance in the sense that we do not discuss quantification as it is common-place in formal semantics. However, cf. chapter 8 of [P&S(1994)] for discussion of a treatment of quantifier scope which could be combined with our approach, if so desired.

A govern*ing* auxiliary will apply argument composition and raise all the complements from the govern*ed* verb(s) to become arguments of the auxiliary, as proposed in [H&N(1989)]. We assume that causative and perceptual verbs syntactically behave just like auxiliaries in this respect.

The difference between auxiliaries on the one hand and perceptual and causative verbs on the other we view as basically semantic. We take it that auxiliaries semantically more or less operate *on* events, affecting features for tense and aspect or modality. Causative and perceptual verbs on the other hand will be analyzed *as* events themselves, events which take other events as their argument, in general as a theme (viz., a value of ARG2, cf. the entry in (7) below).

In chapter 9 of [P&S(1994)] the approach to local selection from [Borsley(1987)] is developed further and leads to the Valence Principle, which refers to the *valence* features SUBJ and COMPS through 'F':

(4) **Valence Principle** Chapter 9, [P&S(1994)]
In a headed phrase, for each valence feature F, the F value of the head-daughter is the concatenation of the phrase's F value with the list of SYNSEM values of the F-daughters' value.

The general effect of the principle on a phrase which is headed by some sign is that this headed sign can only become 'complete' (or "*saturated*") if it is combined with the appropriate arguments. For example, in the case of a transitive verb, such a verb must find a subject NP (selected through SUBJ) and some object (selected through COMPS). If we assume a flat clause structure analysis of Dutch and we furthermore assume lexical signs like (3) and (7), then the immediate dominance statements (5) and (6) will suffice to describe the construction of Dutch we are concerned with here.[7] Here the H,S and C indicate that the daughters of the phrase include a head, a subject and complements, not necessarily in that order (cf. chapter 9 of [P&S(1994)] for details). Note that in addition to the valency features SUBJ and COMPS, we also assume the presence of the GOV-feature, ranging over 1 complement:[8]

(5)  XP[LEX−] → S,C$_1$,..., C$_n$,H[GOV⟨ ⟩, LEX+]

(6)  X[LEX+] → H[GOV ⟨ C$_i$ ⟩, LEX+] , C$_i$

The second schema is in a sense not a "phrase" structure schema but is instead a "cluster-formation"-schema. This is because normally the combination of two or more words leads to a sign which is LEX−, a phrasal sign, but here it leads to a 'complex word' which is LEX+. Also (6) is strictly binary: it takes one argument, namely the argument which is the value of GOV. We arrange the lexicon so that any value of GOV will always be an unsaturated base form verb which is defined as LEX+ as well. By the Valency Principle, this selection requirement of the govern*ing* verb will be appropriately 'cancelled' after string concatenation during parsing.

---

[7] Actually, our analysis also presupposes the Head Feature Principle and Semantics Principle from [P&S(1994)]; cf. Figures 1 and 2 for informal illustration.

[8] Following discussions of Webelhuth, Ackerman, Sag and Pollard at WCCFL XIII, suggesting this for German, and Chung of Ohio State University originally suggesting this for Korean.

Central to our analysis of the case-markings of NPs in the Dutch Mittelfeld is the assumption from [Pollard(fc.)] that base forms of verbs do not assign any case to their subject. The value for the subject-NP's CASE-feature in (3), "case", is the supertype in the type hierarchy for those atomic types that are appropriate values of the feature CASE. So, the value case is the supertype of NOM and ACC in Dutch and English, and in German also of DAT and GEN. The result of assigning the subject-NP this supertype for case in practice boils down to giving this NP some kind of "any"-value for case; the case-value case of such an NP will unify with any other possible case value.

In our analysis, the discontinuous relation between arguments and verbs in DCSDs is brought about firstly by lexically defining finite perceptuals like *zag* (and finite causatives) as *argument composition verbs*, along the following lines:[9]

(7) $\begin{bmatrix} \text{PHON} \langle zag \rangle \\ \text{HEAD} \begin{bmatrix} \text{MAJOR} & \text{v} \\ \text{VFORM} & \text{FIN} \end{bmatrix} \\ \text{SUBJ} \langle \text{NP[NOM]}:\boxed{1} \rangle \\ \text{COMPS} \langle \boxed{5}\text{NP[ACC]} \rangle \oplus \boxed{L} \\ \text{GOV} \langle \begin{bmatrix} \text{V[BASE]} \\ \text{SUBJ} \langle \boxed{5}\text{NP} \rangle \\ \text{COMPS} \boxed{L} \\ \text{CONT} \boxed{4} \\ \text{LEX} + \end{bmatrix} \rangle \\ \text{CONTENT} \begin{bmatrix} \text{DET EVENT} \\ \text{PARA} \boxed{6} \\ \text{RESTR} \left\{ \begin{bmatrix} \text{RELN ZIEN} \\ \text{INST} \boxed{6} \\ \text{ARG1} \boxed{1} \\ \text{ARG2} \boxed{4} \end{bmatrix} \right\} \end{bmatrix} \\ \text{LEX} + \end{bmatrix}$

The finite argument composition verb *zag* selects a singular nominative NP through its SUBJ-feature. As non-subject arguments it selects through its COMPS-feature first the NP tagged as $\boxed{5}$ which is unified with the SUBJ-value of the governed verb(s), and secondly the list $\boxed{L}$ of zero or more non-subject arguments of the governed verb(s). And crucially, being a govern*ing* verb, *zag* selects through GOV a governed base form verb,[10] with as SUBJ-value "$\boxed{5}$", as COMPS-value

---

[9] In this entry and throughout the paper, ⊕ stands for concatenation of arbitrary-length lists of arguments.

[10] One base form verb, or a base form verb-headed verbal clus-

"$\boxed{L}$" and as semantics "$\boxed{4}$". Note that, since the governed V[BSE] is selected *as* missing a subject and a list of complements, it must *not* 'find' this subject or these complements, which it indeed doesn't (cf. the tree in Figure 1).

As it were in passing, the governing perceptual verb (or causative verb alike) imposes accusative case on the NP which denotes the subject-argument of the governed verb. The unification of [CASE case] and [CASE ACC] will be forced through the structure-sharing indicated in (7) as "$\boxed{5}$", and will result in the more specific restriction [CASE ACC]. This accounts for the accusative case-marking on *haar* ("her") in examples (1) and (2), and in general on all non-subject arguments in such constructions.

The second and crucial step in our account of the discontinuity is accounting for the linear order in the verb cluster with DCSDs. The linear order of the verb cluster in Dutch we account for through (8):

(8) Linear Precedence Rule Dutch Verb Clusters
[GOV ⟨ X ⟩] < X

(9) Linear Precedence Rule German Verb Clusters
X < [GOV ⟨ X ⟩]

By these LP-rules, in each part of the binary branching verb cluster the govern*ing* verb will appear head-initial in Dutch, and head-final in German.[11] It is straightforward to show that the above approach has the desired effect also for the sentence (2) mentioned in the introduction if we define a lexical entry for the causative *helpen* with a syntax and semantics along the same lines as the perceptual *zag*. The only difference must be that such nonfinite entries do not assign NOM to their subject, but "case". Other than that, there will just be additional embeddings in the semantics as well as in the verb cluster. Thus, by the ID-rule in (6) and the lexical entries for causatives and perceptuals, we account for the recursiveness of the phenomenon, cf. the tree in Figure 2.

## 4 Conclusion

We extended the [H&N(1989)]-analysis of German to Dutch, accounting for the difference, resp. nested vs. cross serial dependencies, through one single LP-parameter. Also, we argued that such an argument composition approach is to be preferred over several alternative approaches, since argument composition isn't an 'additional' mechanism. Further linguistic advantages of this approach, i.e. accounts of irregular case assignments and constraints on double infinitives, are discussed in [Rentier(1994)]. We are able to derive verb second constructions by standard application of

---

ter; due to the ID-schema in (6) either will be LEX+, so that we are able to recursively build up bigger and bigger LEX+-complexes.

[11] LP-rules like these are common in HPSG, cf. for instance the rule XP < SUBJ ⟨ XP ⟩ which orders subjects before VPs in English ([Borsley(1987)]).

Figure 1: The discontinuous relation: Valence Principle, schema's (5) & (6), entries (3) & (7), LP-rule (8).

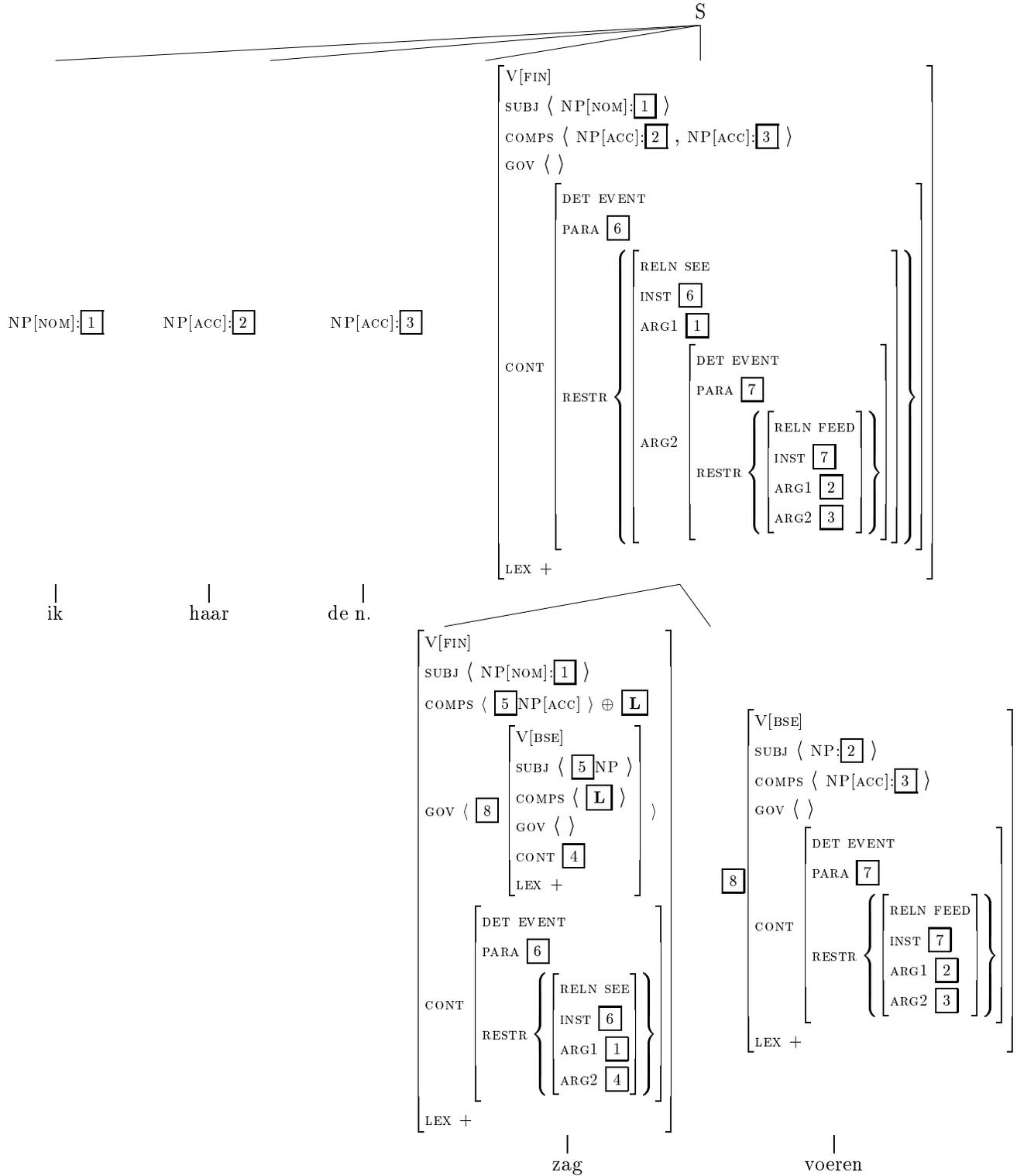

Figure 2: Recursion in the Verb Cluster (Sentence (2)).

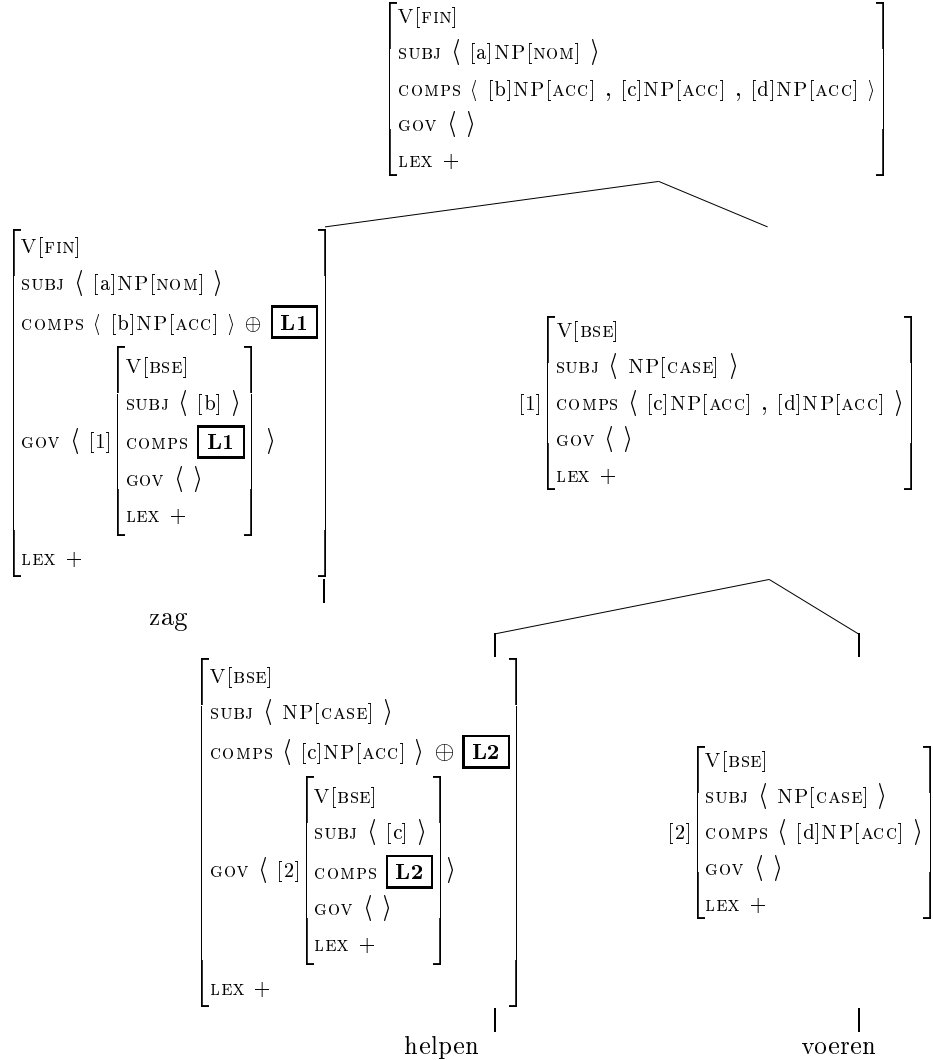

the Dutch versions of the extraction lexical rules (see [Rentier(1993)]) to the verbs at the lexical level.

# References

[Borsley(1987)] R. Borsley, *"Subjects and Complements in HPSG"*, CSLI Report 107, Stanford University, USA

[Davidson(1967)] D. Davidson, "The Logical Form of Action Sentences", reprinted in *"Essays on Actions and Events"*, Clarendon Press, Oxford, 1980

[H&N(1989)] E. Hinrichs, T. Nakazawa, "Flipped out: AUX in German", in *"Proceedings of the 25th Regional Meeting of the Chicago Linguistic Society"*, CLS, USA

[Pollard(fc.)] C. Pollard, "On Head Non-Movement", in: Bunt, H. & van Horck, A. (eds.), *"Proceedings of the Symposium on Discontinuous Constituency"*, Mouton-de Gruyter, Germany

[P&S(1994)] C. Pollard, I.A. Sag, *"Head-driven Phrase Structure Grammar"*, University of Chicago Press and CSLI Publications, USA

[Reape(fc.)] M. Reape, "Getting Things in Order", in: Bunt, H. & van Horck, A. (eds.), *"Proceedings of the Symposium on Discontinuous Constituency"*, Mouton-de Gruyter, Germany

[Rentier(1993)] G. Rentier, "Dutch Object Clitics, Preposition Stranding and Across-the-Board Extraction", in Sijtsma, W. & Zweekhorst, O. (eds.), *"Papers from Computational Linguistics in the Netherlands (CLIN) III, 1992"*, the Netherlands

[Rentier(1994)] G. Rentier "A Lexicalist Approach to Dutch Cross Serial Dependencies" in: *"Proceedings of the 30th Regional Meeting of the Chicago Linguistic Society"*, Chicago, CLS USA

[Rentier(ms.)] G. Rentier, *"Head-driven Phrase Structure Grammar and Underspecified Logical Form"*, ms., ITK, Tilburg University, the Netherlands